\documentclass[preprint,12pt]{elsarticle}

\usepackage{amssymb}
\usepackage{amsmath}
\usepackage{graphicx}
\usepackage{booktabs}
\usepackage{multirow}
\usepackage{bm}
\usepackage{adjustbox}

\journal{Information Fusion}

\begin{document}

\begin{frontmatter}



\title{A brain-inspired information fusion method for enhancing robot GPS outages navigation}


\author[label1]{Yaohua Liu \corref{cor1}} 
\author[label2]{Hengjun Zhang}
\author[label3]{Binkai Ou} %% Author name

\affiliation[label1]{organization={Guangdong Institute of Intelligence Science and Technology},%Department and Organization
            addressline={Hengqin}, 
            city={Zhuhai},
            postcode={519031}, 
            state={Guangdong},
            country={China}}
\affiliation[label2]{organization={School of Electronic Engineering and Automation, Guilin University of Electronic Technology},%Department and Organization
            city={Guilin},
            postcode={541000}, 
            state={Guangxi},
            country={China}}            
\affiliation[label3]{organization={Innovation and Research and Development Department,  BoardWare Information System Company Ltd},%Department and Organization
% addressline={Beijing}, 
city={Macau},
postcode={999078}, 
% state={Beijing},
country={China}}
            
\cortext[cor1]{Corresponding author} 

%% Abstract
\begin{abstract}
Low-cost inertial navigation systems (INS) are prone to sensor biases and measurement noise, which lead to rapid degradation of navigation accuracy during global positioning system (GPS) outages. To address this challenge and improve positioning continuity in GPS-denied environments, this paper proposes a brain-inspired GPS/INS fusion network (BGFN) based on spiking neural networks (SNNs). The BGFN architecture integrates a spiking Transformer with a spiking encoder to simultaneously extract spatial features from inertial measurement unit (IMU) signals and capture their temporal dynamics. By modeling the relationship between vehicle attitude, specific force, angular rate, and GPS-derived position increments, the network leverages both current and historical IMU data to estimate vehicle motion. The effectiveness of the proposed method is evaluated through real-world field tests and experiments on public datasets. Compared to conventional deep learning approaches, the results demonstrate that BGFN achieves higher accuracy and enhanced reliability in navigation performance, particularly under prolonged GPS outages.
\end{abstract}


% \begin{highlights}
% \end{highlights}

%% Keywords

\end{frontmatter}



\section{Introduction}
\label{sec1}
%% Labels are used to cross-reference an item using \ref command.

  The integration of the inertial navigation system (INS) and global positioning system (GPS) provides a high-precision navigation solution that is widely used in ground vehicles and unmanned aerial vehicles (UAVs) when GPS signals are available \cite{wang2019constrained}. However, when these vehicles travel through environments with weak GPS signals, such as areas surrounded by tall buildings or inside tunnels, the GPS reception is often blocked. As a result, the integrated system operates in standalone INS mode. In this case, navigation accuracy deteriorates rapidly due to errors in the inertial measurement unit (IMU), including bias drift and scale factor instability \cite{kim2017robust}. Therefore, it is essential to develop integrated navigation systems that can maintain reliable performance in a variety of complex and challenging environments.

To mitigate the adverse effects of GPS signal interruptions, a range of fusion strategies have been proposed for enhancing GPS/INS navigation performance. These can generally be categorized into two main approaches. The first involves incorporating auxiliary sensing technologies—such as lidar \cite{wang2025recent}, vision systems \cite{tabassum2025comparative}, or wheel odometry \cite{li2024resilient}—to compensate for the absence of GPS updates. While effective in improving accuracy, this approach increases both the system’s cost and its structural complexity. The second relies on machine learning (ML) techniques, including ensemble learning \cite{li2017improving}, support vector machines (SVMs) \cite{bhatt2014novel}, and random forest regression (RFR) \cite{adusumilli2013low}, to emulate the error propagation behavior of standalone INS. By leveraging datasets where GPS measurements are available, these models learn the relationship between vehicle dynamics and INS error states, enabling error prediction during GPS outages.

The advancement of artificial intelligence (AI) has led to widespread use of artificial neural networks (ANNs) for improving navigation accuracy. Early studies, such as those by Rashad et al. \cite{sharaf2005online}, employed radial basis function neural networks (RBFNNs) to map the relationship between INS-derived positions and the associated errors using GPS data. Similarly, El-Sheimy et al. \cite{el2006utilization} developed an ANN-based INS/GPS integration approach to fuse uncompensated INS outputs with differential GPS measurements. Later work by Zhang et al. \cite{zhang2018hybrid} introduced a dual optimization framework that combined cubature Kalman filtering (CKF) with multilayer perceptron (MLP) networks, enhancing model precision by adaptively tuning network weights. Doostdar et al. \cite{doostdar2020ins} further explored recurrent fuzzy wavelet neural networks (RFWNNs) to model the connection between INS motion parameters and GNSS/INS discrepancies, enabling more precise corrections during GNSS outages. More recently, Belhajem et al. \cite{belhajem2018improving} proposed robust INS/GPS fusion methods integrating extended Kalman filters (EKFs) with ML algorithms such as neural networks and SVMs, yielding improved stability and accuracy.

Despite their promise, most of these approaches are based on static neural network architectures that rely only on current and immediate past INS data, without leveraging a broader history of vehicle dynamics. The absence of long-term temporal context limits localization accuracy during extended GPS interruptions. In recent years, deep learning (DL) techniques have shown superior performance in time-series prediction tasks, including natural language processing and speech recognition
\cite{li2024unified}. Recurrent neural networks (RNNs) are particularly well-suited to modeling sequential data and nonlinear system behavior, offering advantages over traditional ML methods \cite{liu2021vehicle,zhao2022novel,xu2022motion,fang2020lstm}. For example, RNNs combined with unscented Kalman filters (UKFs) have been used to estimate and compensate for MEMS gyroscope drift in real time. However, conventional RNNs often suffer from vanishing or exploding gradients when handling long sequences. Long short-term memory (LSTM) networks have been developed to address these issues, enabling denoising of MEMS IMU signals and improved estimation of INS errors using both present and past outputs \cite{li2021recurrent}, \cite{chen2024deep}. Hybrid architectures combining convolutional neural networks (CNNs) with gated recurrent units (GRUs) have also been explored for extracting IMU features from noisy measurements \cite{liu2022deep}.

While these DL-based methods demonstrate notable improvements, their dependence on CNNs and RNNs constrains their ability to capture long-term dependencies and extract motion-relevant patterns from IMU sequences heavily affected by random noise. Transformer-based models, such as the bidirectional encoder proposed by Guyard et al. \cite{guyard2025transformer}, have shown potential in modeling extended temporal patterns for INS/GPS fusion during GPS outages. Nonetheless, these methods typically require large, high-quality datasets to generalize effectively, which limits their applicability in low-data scenarios.

Spiking neural networks (SNNs), regarded as the third generation of neural network models, have emerged as a promising alternative. SNNs can match the performance of conventional DL models in various time-series applications while offering greater computational efficiency and robustness in temporal feature extraction \cite{lv2024efficient,candelori2025spatio,li2025neurove}. Their event-driven processing paradigm enables effective handling of low-cost MEMS IMU data with minimal energy consumption, making them well-suited for embedded and resource-constrained navigation systems.

This paper proposes an advanced fusion strategy that integrates the Kalman Filter (KF) with a novel brain-inspired GPS/INS fusion network (BGFN) to maintain high navigation accuracy during periods of GPS signal loss. Within the BGFN framework, a spiking encoder is utilized to transform IMU inputs into a high-dimensional spiking representation, facilitating more effective extraction of IMU features from noisy sensor data. Additionally, a spiking Transformer architecture is introduced to capture and model the dynamic motion patterns of mobile robots over time. To support the INS during GPS outages while minimizing computational overhead, a hybrid fusion mechanism leveraging the spiking Transformer is designed. The key contributions of this work are summarized as follows:

\begin{enumerate}
  \item A novel brain-inspired GPS/INS fusion network, termed BGFN, is introduced to enhance the accuracy of integrated GPS/INS navigation. To the best of the authors’ knowledge, this work represents the first application of a spiking Transformer in the context of GPS/INS fusion. Unlike conventional deep learning-based approaches that primarily rely on current sensor inputs, the proposed hybrid fusion strategy within BGFN incorporates both real-time INS errors and historical inertial data. This enables more accurate modeling of INS error evolution and improved estimation of vehicle dynamic states, particularly during extended GPS outages.
  \item A hybrid fusion strategy is proposed to model the nonlinear relationship between sensor measurements and GPS position increments during periods of GPS signal loss. Specifically, during periods of GPS availability, the KF is employed to fuse INS and GPS data, yielding accurate and reliable position and velocity estimates. Concurrently, the synchronized INS and GPS measurements are stored onboard for training the BGFN model. When GPS signals become unavailable, the pre-trained BGFN is activated to predict the GPS position increments, which are then integrated into the navigation solution to sustain high positioning accuracy.
  \item In comparison to conventional neural network models, the proposed BGFN demonstrates superior capability in extracting meaningful spatial features from IMU data amidst sensor noise, thereby improving measurement accuracy. Moreover, thanks to its spiking neural network architecture, BGFN achieves these enhancements with significantly lower computational energy consumption. The performance of the proposed method is comprehensively evaluated through a series of experiments using publicly available datasets as well as real-world field tests, demonstrating its effectiveness and robustness in practical navigation scenarios.
\end{enumerate}

The remainder of the paper is organized as follows. In Section II, the mathematical model of the GPS/INS integrated system and KF is explained in detail. Section III describes a process of establishing a navigation model based on BGFN in GPS denied scenarios. Dataset experiments and real field tests are performed and discussed in Section IV. The conclusion is provided in Section V.

\section{GPS/INS Integrated Navigation System Model}
\subsection{INS Error Model}
The INS serves as a fundamental component in both INS/GPS and INS/ML integration frameworks, delivering estimates of a vehicle’s attitude, velocity, and position by double-integrating inertial sensor data from the IMU. However, due to inherent sensor biases and cumulative integration errors, the position and velocity solutions from standalone INS tend to drift rapidly over time. To mitigate this limitation, INS is commonly fused with GPS using a KF to improve overall navigation accuracy and reliability. The architecture of GPS/INS integration can be categorized into three main configurations: loosely coupled, tightly coupled, and ultra-tightly coupled. Among these, loosely coupled integration is widely adopted in vehicle and UAV navigation applications due to its simplicity, ease of implementation, and robust performance under normal operating conditions.

\begin{figure}[!htp]
  \centering
  \includegraphics[width=.95\linewidth]{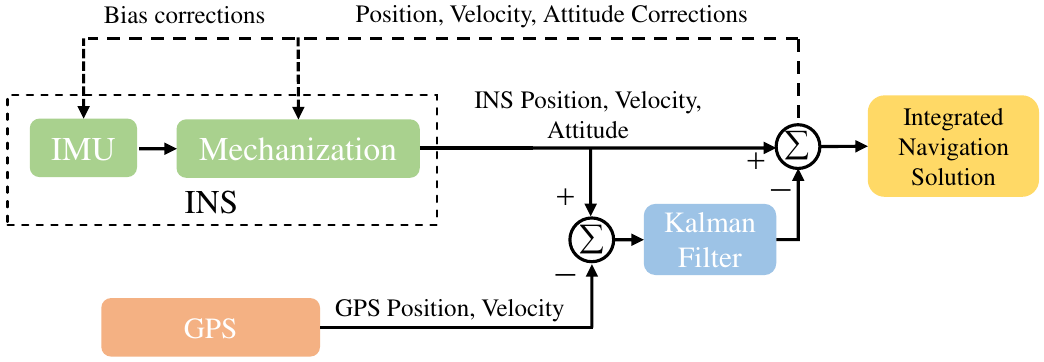}\\
  \caption{A block diagram of a loosely coupled GPS/INS integration.}\label{ins-gps-loose}
\end{figure}

A loosely coupled GPS/INS integrated navigation system is illustrated in Fig. \ref{ins-gps-loose}. In this configuration, GPS and INS operate independently, each generating its own navigation solution. To enhance accuracy, the position and velocity outputs from both systems are fed into a KF, which computes their differences and estimates the INS errors based on a predefined error model. These estimated errors are then used to correct the INS solution, resulting in a refined, integrated navigation output that provides improved estimates of position, velocity, and attitude.

Accurate navigation for vehicles and UAVs relies on the proper selection of coordinate systems, particularly the body-fixed frame and the navigation frame. The navigation frame is typically defined as the local geographic coordinate system, which aligns with the north, east, and down (NED) directions and follows the right-hand coordinate rule. This reference frame is widely used in inertial navigation due to its intuitive alignment with geographic directions. A comprehensive derivation of the dynamic error model for the INS is presented in the next section. By omitting higher-order small terms and certain sensor non-idealities, the attitude error equation of the INS can be formulated as,
\begin{equation}\label{eq9}
\dot \phi  = \phi  \times \omega _{in}^n + \delta \omega _{in}^n - {\varepsilon ^n}
\end{equation}
where $\phi$ is the attitude angle error; $\omega _{in}^n$ and $\delta \omega _{in}^n$ are the angular velocity of the rotation of the navigation coordinate frame relative to the inertial coordinate frame and its error; $\varepsilon ^n$ is the gyroscope drift vector of the navigation coordinate frame.

The INS velocity error equation is described as,
\begin{equation}\label{IVE}
\begin{array}{c}
\delta {\dot V^n} =  - {\phi ^n} \times {f^n} + \delta {V^n} \times (2\omega _{ie}^n + \omega _{en}^n) + {V^n}\\
 \times (2\delta \omega _{ie}^n + \delta \omega _{en}^n) + {\nabla ^n}\\
\end{array}
\end{equation}
where $\delta {V^n}$ and $V^n$ are the velocity error and velocity in the east, north and upward, respectively; $f^n$ is the specific force; $\omega _{ie}^n$ and $\omega _{en}^n$ are the earth self-rotation rate and the angle rate relative to the earth in the navigation coordinate system, respectively; ${\nabla ^n}$ is the accelerometers' bias of the navigation frame.

The INS position error equation can be expressed as,
\begin{equation}\label{IPE}
\left\{
\begin{array}{l}
\delta \dot L = \frac{{\delta {V_N}}}{{{R_M} + h}} - \frac{{\delta h{V_N}}}{{{{({R_M} + h)}^2}}}\\
\delta \dot \lambda  = \frac{{\delta {V_E}\sec L}}{{{R_N} + h}} + \frac{{\delta L{V_E}\tan L\sec L}}{{{R_N} + h}} - \frac{{\delta h{V_E}\sec L}}{{{{({R_N} + h)}^2}}}\\
\delta \dot h = \delta {V_U}
\end{array}
\right.
\end{equation}
where $\delta L$, $\delta \lambda$ and $\delta h$ are the errors of latitude, longitude and height; $\delta V_N$, $\delta V_U$ and $\delta V_E$ denote the velocity errors in north, east and upward directions respectively; $R_M$ and $R_N$ are the radiuses of the curvature in the meridian and prime vertical.

\subsection{The Information fusion model based on Kalman filter}
The KF is a well-established algorithm in control theory, also referred to as a linear quadratic estimator (LQE), renowned for its ability to estimate unobserved system states with higher accuracy than methods relying on individual measurements \cite{peng2022srai},\cite{zhang2019time}. To apply the KF for state estimation, the system should first be described by a set of state and measurement equations. Assuming that the true system state at the ${{\rm{k}}^{th}}$ time step evolves from the state at step $k-1$, the discrete-time state transition can be expressed as,
\begin{equation}\label{eq1}
{x_k} = {F_k}{x_{k - 1}} + {B_k}{u_k} + {w_k}
\end{equation}
where $x_k$ is the state vector at the ${{\rm{k}}^{th}}$ moment; $F_k$ is the state transition model applied to the previous state $x_{k-1}$; $B_k$  is the coefficient vector related to the control vector $u_k$; $w_k$ is the process noise assumed as a zero mean with normal distribution and covariance matrix $Q_k$.

The measurement equation can be defined as,
\begin{equation}\label{eq2}
{z_k} = {H_k}{x_k} + {v_k}
\end{equation}
where $z_k$ is the measurement vector; $H_k$ is the observation model which maps the actual state space into the observed space; $v_k$ is the observation noise assumed to be zero mean Gaussian white noise with covariance $R_k$.

If the estimated state vector $x_k$ and measurement vector $z_k$ can be written in the form of Eq. \ref{eq1} and Eq. \ref{eq2}, and the noise variance $w_k$ and $v_k$ follow the zero mean Gaussian white noise distribution, the process of KF algorithm can be divided into two parts:

\noindent{ (i) Time update: }

\begin{equation}\label{eq3}
% \setstretch{2.0}
{\begin{array}{*{20}{c}}
  \hat x_k^ -  = {F_k}{\hat x_{k - 1}} + {B_k}{u_k} \\
  P_k^ -  = {F_k}{P_{k - 1}}{F_k^T} + Q  \\
\end{array}}
\end{equation}

\noindent{ (ii) Measurement update: }

\begin{equation}\label{eq4}
% \setstretch{2.0}
{\begin{array}{*{20}{c}}
  {K_k} = P_k^ - {H_k^T}{(H_kP_k^ - {H_k^T} + R)^{ - 1}}   \\
 {\hat x_k} = \hat x_k^ -  + {K_k}({z_k} - H_k\hat x_k^ - )   \\
 {P_k} = (I - {K_k}H_k)P_k^ - \\
\end{array}}
\end{equation}

In Eq. \ref{eq3}, $\hat x_k^ -$ denotes the a priori estimate of the system state at time step $k$, and $P_k^ -$ represents the corresponding a priori error covariance matrix. In Eq. \ref{eq4}, $P_k$ is the updated (a posteriori) state covariance matrix, and $K_k$ is the Kalman gain matrix, which determines the weight given to the measurement residual in correcting the state estimate. The core mechanism of the KF involves the iterative execution of two phases: prediction (time update) and correction (measurement update). Through this recursive process, the optimal state estimate $\hat x_k$ is computed at each time step, provided that initial values for the state vector $x_0$ and the error covariance matrix $P_0$ are available.

The INS error model in Section II is suitable for vehicle stop and motion. The KF with a 15-state vector $\textbf{x}$ can be employed to correct the INS errors, and the system state vector $\textbf{x}$ is defined as,
\begin{equation}\label{eq13}
\begin{array}{*{20}{c}}
{\textbf{x}}=[{\varphi _E}&{\varphi _N}&{\varphi _D}&{\delta {V_E}}&{\delta {V_N}}&{\delta {V_D}}&{\delta L}&{\delta \lambda }\\
& {\delta h} & {\nabla _x} & {\nabla _y} & {\nabla _z} & {\varepsilon _x} & {\varepsilon _y} & {\varepsilon _z}]
\end{array}
\end{equation}
where $({\varphi _E},{\varphi _N},{\varphi _D})$ are the misalignment angles of the calculated platform in the east-north-up navigation coordinate system; $(\delta {V_N},\delta {V_E},\delta {V_D})$ are the velocity errors. $\delta L$, $\delta \lambda$ and $\delta h$ are the latitude, longitude and height errors, respectively. ${\nabla _{{{\rm{x}}_,}{y_,}z}}$ and ${\varepsilon _{{{\rm{x}}_,}{y_,}z}}$ denote the accelerometer biases and gyro biases in the body coordinate system.

The measurement vector $z$ is defined as the difference in position between the INS and GPS solutions. Under normal operating conditions, the GPS/INS integrated system operates continuously by fusing data through the KF, as governed by the previously described equations. However, during GPS outages, the KF loses access to external position updates, preventing it from effectively estimating and correcting the accumulating INS errors. As a result, the navigation solution becomes increasingly degraded over time due to unbounded error growth in the standalone INS.

\subsection{Spiking Neurons and Surrogate Gradient}
\begin{figure}[!ht]
    \centering
    \includegraphics[width=.9\linewidth]{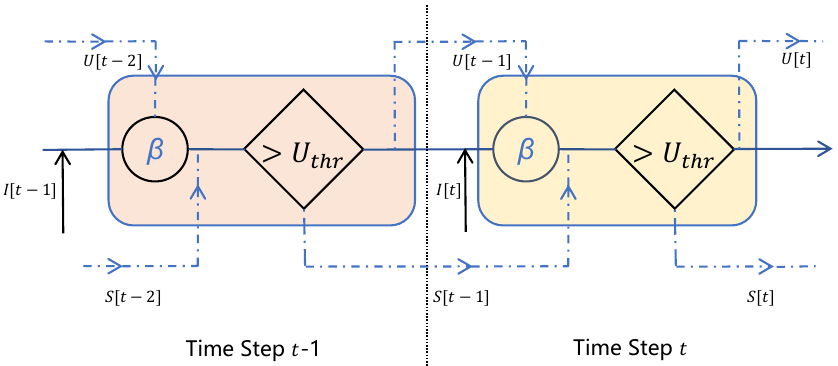}
    \caption{The structure illustration of the LIF neuron.}
    \label{fig:LIF}
  \end{figure}

In contrast to conventional deep neural networks (DNNs), the proposed method employs SNNs whose basic computational unit is the leaky integrate-and-fire (LIF) neuron \cite{maass1997networks}, a model that more closely mimics biological neuronal behavior. As depicted in Fig. \ref{fig:LIF}, the LIF neuron accumulates incoming synaptic inputs over time in the form of spikes. When the membrane potential reaches a predefined threshold, the neuron generates an output spike and resets its potential. Between spikes, the membrane potential gradually decays, reflecting the "leaky" nature of the integration process. The dynamics of the membrane potential can be formally expressed as,
\begin{equation}
\begin{array}{l}
  U(t)=H(t-\Delta t)+I(t), \quad I(t)=f(x ; \theta),                                                                         \\
  H(t)=V_{\text{reset }} S(t)+(1-S(t)) \beta U(t),                                                                                    \\
  S(t)=\left\{\begin{array}{ll}1, & \text{ if } U(t) \geq U_{\text{thr }}, \\ 0, & \text{ if } U(t)<U_{\text{thr }}\end{array}\right.
\end{array}
\label{eq:LIF}
\end{equation}

In the context of the LIF neuron model, the spatial input $I(t)$ at time step $t$ is computed by applying a learnable function $f$ to the input $\bf{x}$, parameterized by $\theta$. The temporal output $H(t)$ is governed by a discretization constant $\Delta t$, which controls the resolution of the LIF dynamics. The spike response $S(t)$ is defined using a Heaviside step function, triggered when the membrane potential $U(t)$ exceeds a predefined threshold $ U_{\text{thr }}$. Upon firing, the neuron emits a spike and the membrane potential is reset to $V_{reset}$. If the threshold is not reached, no spike is generated, and the membrane potential decays toward $H(t)$ at a rate determined by the decay rate $\beta$. This formulation enables the neuron to capture temporal dependencies and sparse activation patterns, which are crucial for efficient spatiotemporal feature extraction.

Within the LIF neuron framework, the spatial input $I(t)$ at time step $t$ is obtained by applying a learnable mapping ${f_\theta }$ to the input vector $x$, where $\theta$ represents the trainable parameters. The temporal dynamics of the neuron are discretized using a time step $\Delta t$, which determines the resolution of the simulation. The spiking behavior $S(t)$ is modeled via a Heaviside step function, activated when the membrane potential $U(t)$ exceeds a fixed threshold $ U_{\text{thr }}$. Upon spiking, a pulse is emitted and $U(t)$ is reset to $V_{reset}$. In the absence of a spike, the potential gradually decays toward the resting state $H(t)$, governed by a decay factor $\beta$. This dynamic enables the LIF neuron to effectively capture temporal correlations and sparse, event-driven activation patterns—key characteristics for efficient spatiotemporal representation learning in spiking neural networks.

Through a spiking neuron layer $SN( \cdot )$, the spike trains $S$ are generated by iterating over ${T'}$ discrete time steps, where each of the $N$ input currents $I$ is processed by a corresponding LIF neuron. Formally, this can be expressed as,
\begin{equation}
{S}= SN(I)\label{eq:spike_layer}
\end{equation}
where the arctangent-like surrogate gradients \cite{fang2023spikingjelly} are used
to approximate the gradients of the spiking neurons during the backpropagation
process. The surrogate gradient is defined as follows,
\begin{equation}
S(t) \approx \frac{1}{\pi }\arctan (\frac{\pi }{2}\alpha U(t)) + \frac{1}{2}\label{eq:surrogate_gradient}
\end{equation}
where $\alpha$ is a hyperparameter that controls the steepness of the surrogate
gradient. Therefore, the comprehensive model can be trained using an end-to-end
approach, facilitated by backpropagation through time (BPTT) \cite{gruslys2016memory}, which enables efficient
learning and optimization.
  
\section{A Brain-inspired GPS/INS fusion Method}
\subsection{The hybrid GPS/INS navigation structure}
The proposed GPS/INS integrated navigation system employs an SNN to model the relationship between navigation information and the outputs of the INS, including velocity, attitude, specific force, and angular rate. This approach maintains high navigation accuracy even during GPS outages. Navigation information generally refers to either the position difference between the INS and GPS solutions or the GPS position increment. The main motivation for using SNNs stems from the highly nonlinear and complex nature of the GPS/INS system, which is difficult to describe with a precise mathematical model. Traditional filter-based methods depend heavily on accurate system modeling, and their performance deteriorates when sensor data is missing or degraded, such as during GPS signal loss. In contrast, SNN-based methods are capable of capturing temporal patterns and extracting meaningful spatial features from noisy measurements. Compared to conventional machine learning-aided approaches, the SNN method demonstrates enhanced robustness and lower energy consumption, making it particularly suitable for navigation in challenging environments where GPS signals are unavailable.

Recently, numerous ML-aided models have been proposed to capture the relationship between navigation information and INS outputs, which can generally be classified into three categories based on their target variables:  $O_{INS}-\delta P_{INS}$, $O_{INS}-X_k$ and $O_{INS}-\Delta P_{GPS}$. The $O_{INS}-\delta P_{INS}$ model aims to learn the relationship between INS measurements and the position error of the integrated GPS/INS system, effectively estimating the deviation of the INS solution from the GPS reference. The $O_{INS}-X_k$ approach focuses on establishing a mapping from INS data to the Kalman filter state vector $X_k$, which typically includes navigation errors and sensor biases. In contrast, the $O_{INS}-\Delta P_{GPS}$ model uses INS outputs as input and predicts the GPS-derived position increment over a given time interval. Notably, the first two models incorporate both INS and GPS information in their target outputs, which may introduce coupling errors due to GPS noise or filter artifacts. In comparison, the $($$O_{INS}-\Delta P_{GPS}$$)$ model relies solely on short-term GPS position changes, reducing dependency on long-term GPS accuracy and minimizing error propagation, thereby offering greater robustness during GPS outages. The GPS position increment in this model can be expressed as \cite{yao2017rls},
\begin{equation}\label{eq8-1}
\begin{array}{l}
\begin{array}{l}
\Delta {P_{GPS}} = \iint{{{\dot V}_n}(t)}dtdt
\end{array}\\
=\begin{array}{l}
\iint{(C_b^nf_{ib}^b(t)-(2\omega _{ie}^n(t)+\omega _{en}^n(t))\times {V_n}(t)+G_n)dtdt}
\end{array}
\end{array}
\end{equation}
where $C_b^n$ is the direction cosine matrix of the transformation from body frame to the navigation frame; $f_{ib}$ is the output of the accelerometer; $\omega _{ie}^n$ is the angular rate of earth frame $e$ to the inertial frame $i$; $\omega _{en}^n$ is the angular rate of the navigation frame $n$ to the earth frame $e$; $V_n$ is the velocity of the vehicle in the navigation frame $n$ and $G_n$ is the gravity vector.

In Eq. $($\ref{eq8-1}$)$, the terms $\omega _{ie}^n$ and $G_n$ depend on the longitude and latitude, while $\omega _{en}^n$ is a function of the velocity vector $V_n$. However, in practical navigation scenarios, the variations in longitude and latitude are typically small over short time intervals. As a result, the dominant factors influencing the GPS position increment $\Delta P_{GPS}$ are the attitude matrix $C_b^n$, the specific force measurements $f_{ib}$, and the velocity vector $V_n$. Among these, the transformation matrix  $C_b^n$, which maps vectors from the body frame to the navigation frame, plays a critical role and can be expressed as,
\begin{equation}\label{cnb}
\scriptsize
\left[ {\begin{array}{*{20}{c}}
{\cos \theta \cos \psi }&{ - \cos \gamma \sin \psi  + \sin \gamma \sin \theta \cos \psi }&{\sin \gamma \sin \psi  + \cos \gamma \sin \theta \cos \psi }\\
{\cos \theta \sin \psi }&{\cos \gamma \cos \psi  + \sin \gamma \sin \theta \sin \psi }&{ - \sin \gamma \cos \psi  + \cos \gamma \sin \theta \sin \psi }\\
{ - \sin \theta }&{\sin \gamma \cos \theta }&{\cos \gamma \cos \theta }
\end{array}} \right]
\end{equation}
where $\theta$, $\gamma$ and $\psi$ are the pitch, roll and yaw, respectively. The attitude angles are mostly obtained by integrating the gyroscope outputs $\omega _{ib}^b$. In summary, $\Delta P_{GPS}$ is mainly determined by $f_{ib}$,  $\omega _{ib}^b$, $\theta$, $\gamma$ and $\psi$, which are selected as the inputs of BGFN model to mimic the mathematical relationship of the position increments of the GPS/INS integrated system when GPS signals are available.

The output is the GPS position increments $\Delta P_{GPS}$, and the input and output can be expressed as,
\begin{equation}\label{eq21}
\begin{array}{l}
Input:[\begin{array}{*{20}{c}}
{\phi }&{{f^b}}&{{\omega ^b}}
\end{array}]\\
 = [\begin{array}{*{20}{c}}
{\gamma }&{\theta }&{\psi}&{{f_x^b}}&{{f_y^b}}&{{f_z^b}}&{{\omega_x^b}}&{{\omega_y^b}}&{{\omega_z^b}}
\end{array}]\\
Output:[\begin{array}{*{20}{c}}
{\Delta P_{GPS}}
\end{array}] = [\begin{array}{*{20}{c}}
{\Delta {P_{GPS}^n}}&{\Delta {P_{GPS}^e}}
\end{array}]
\end{array}
\end{equation}

\begin{figure}[!htp]
  \centering
  \includegraphics[width=.9\linewidth]{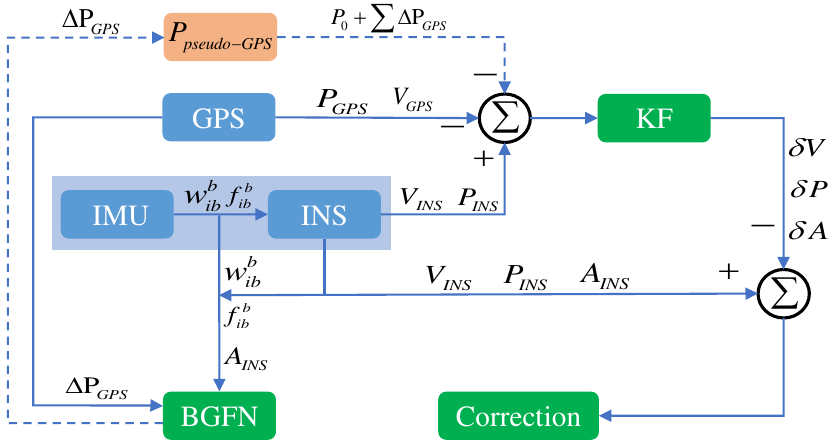}\\
  \caption{The GPS/INS integrated navigation system based on BGFN.}\label{BGFN}
\end{figure}

At the initial stage, the pre-trained BGFN is deployed on the mobile vehicle. Once the vehicle begins motion, the GPS/INS integrated navigation system based on BGFN operates in two distinct modes. As illustrated in Fig. \ref{BGFN}, when GPS signals are available, the system enters the online-training mode. In this mode, the INS-provided velocity $V_{INS}$, position $P_{INS}$, and attitude $A_{INS}$ are fused with the GPS position $P_{GPS}$ within a KF. The KF estimates the velocity error $\delta V$, position error $\delta P$, and attitude error $\delta A$, which are then fed back to correct the INS solutions and mitigate position drift. Concurrently, the synchronized GPS/INS data are logged on the onboard computer to further train the BGFN. Since GPS outages typically occupy only a small fraction of the total operation time, the BGFN remains in training mode for most of the mission. During this extended period, IMU data are continuously fed into the network, enabling the BGFN to accumulate sufficient representative data for comprehensive training. Through this process, the synaptic weights in the hidden layers are iteratively adjusted to accurately capture the underlying input-output dynamics between inertial measurements and navigation increments.

Once GPS signals become unavailable, the BGFN-based GPS/INS navigation system transitions into prediction mode. During GPS outages, no new external position updates are available, and the KF estimates are held at their last valid values, preventing further correction of INS errors. As a result, the system effectively operates as a standalone INS, which is prone to accumulating errors over time. The architecture of this prediction mode is illustrated by the dashed-line components in Fig. \ref{BGFN}. In this mode, the pre-trained BGFN predicts the GPS position increments $\Delta P_{GPS}$, which are then accumulated to generate a pseudo-GPS position. This synthetic position solution is computed by integrating all predicted increments according to Eq. \ref{pred-gps}.
\begin{equation}
P_{GPS}(k) = P_{GPS_0} + \mathop \sum \limits_{i = 0}^k \Delta P_{GPS_i}
\label{pred-gps}
\end{equation}
where $P_{GPS_0}$ is the initial position when GPS fails at the ${\rm{k}}^{th}$ moment. This estimated position $P_{GPS}(k)$ is then used instead of the missing GPS position increments and fused with INS by KF. The hybrid structure of GPS/INS will maintain navigation continuously when GPS signals are lost.

\subsection{The BGFN Architecutre}
\begin{figure}[!htp]
  \centering
  \includegraphics[width=.95\linewidth]{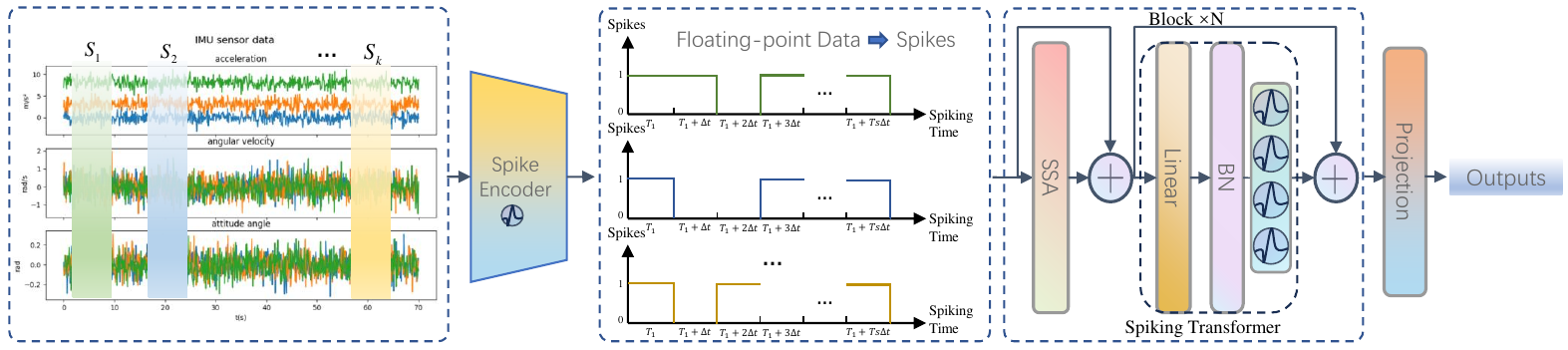}\\
  \caption{The architecture of the BGFN.}\label{BGFN-arc}
\end{figure}

As illustrated in Fig. \ref{BGFN-arc}, the BGFN consists primarily of spike encoder layers and spiking Transformer layers. The spike encoder aligns the temporal resolution between the IMU time-series data and the spiking neural network while converting the continuous sensor measurements into meaningful spike trains. The spiking Transformer layers then process these spike sequences to capture the dynamic motion patterns of the mobile vehicle from historical IMU data. The input to the network is represented as a time series vector ${S_t} = ({u_1},{u_2}, \ldots, {u_i}, \ldots {u_k})$,  where each ${u_i}$ contains nine-dimensional measurements from the accelerometer, gyroscope, and attitude sensors at time step $i$, and $k$ denotes the size of the time window. The network predicts the GPS position increment $\Delta P_{GPS}^{k{\rm{ + }}1}$ at the next time step ${k+1}$ based on the preceding $k$ IMU observations. Thus, the navigation-aiding prediction task can be formulated as, 

% \begin{equation}\label{eq17_1}
% \Delta P_{GPS}^{k{\rm{ + }}1} = GI{\rm{-}}NN({S_1},{S_2}, \ldots ,{S_k})
% \end{equation}
 \begin{equation}
    \Delta P_{GPS}^{k{\rm{ + }}1}= BGFN({S_1},{S_2}, \ldots ,{S_k})\label{eq:bgfn_net}
  \end{equation}

To fully exploit the temporal processing capabilities of SNNs, it is crucial to align the time scale of the IMU time-series data with the dynamics of the SNN. As shown in Fig. \ref{BGFN-arc}, the key strategy in this work involves capturing fine-grained temporal patterns of spike activity within the sensor data at each time step. To achieve this alignment, each time step $\Delta T$ of the input sequence is divided into $T_s$ discrete segments, each of duration $\Delta t$, such that $\Delta T ={T_s}\Delta t$. This relationship establishes a direct correspondence between the time resolution of the IMU data and the internal time step of the SNN. As a result, the independent variable $t$, which is used in both the time-series signal $X(t)$ and the SNN state variables such as membrane potential $U(t)$, input current $I(t)$, hidden state $H(t)$ and spike activity $S(t)$, acquires a unified temporal meaning across the network. Inspired by the recent work of Qu et al. \cite{qu2024cnn}, which demonstrated that such temporal patterns can be effectively captured using convolutional kernels, a one-dimensional convolutional layer is employed as a temporal encoder to transform raw IMU inputs into spike-compatible representations. Given the $k_{th}$ historical IMU sequence ${S_t}$, it is passed through the convolutional layer followed by batch normalization to generate the corresponding spike trains as,
  \begin{equation}
    {S}= SN(BN(Conv(S_k)))\label{eq:spike_input}
  \end{equation}

By passing through the convolutional-based spike encoder, the dimension of the input $S_k$ is expanded to  ${T_s}\times T \times C$, where $C$ denotes the number of sensor channels, specifically nine, corresponding to the three-axis angular velocity, three-axis linear acceleration, and three attitude angles. The convolutional spike encoder effectively captures the intrinsic temporal patterns within the input data, including dynamic changes in magnitude and waveform shape over time. This rich temporal representation enhances the robustness of the encoded features and accurately reflects the time-varying nature of the IMU signals. As a result, the encoded spike tensor is well-suited for subsequent spiking layers, enabling efficient event-driven processing and facilitating accurate modeling of motion dynamics in the spiking neural network.

Given the high sampling frequency of the IMU, typically 100 Hz, and the presence of substantial nonlinear noise in the sensor data, the spiking neural network need learn temporal features over long time windows. Traditional neural networks such as RNNs and CNNs often struggle to capture long-term dependencies due to the vanishing gradient problem. To address this challenge, we employ a spiking Transformer as the backbone of the SNN, which is capable of modeling long-range temporal dependencies and extracting discriminative temporal patterns from IMU measurements. In the proposed architecture, the spiking Transformer is adapted from SpikeFormer v2 \cite{zhou2024spikformer}, which has achieved state-of-the-art performance on large-scale vision benchmarks including ImageNet-1K and CIFAR-10. This strong representational capability motivates its application to inertial navigation, where accurate modeling of dynamic motion over time is critical.

Inspired by Spikformer v2, the spiking self-attention (SSA) mechanism is employed to construct the spiking Transformer blocks. The SSA mechanism is specifically designed to capture temporal dependencies and long-range relationships within the input spike sequences, which is crucial for accurately modeling the dynamic characteristics of IMU measurements under high-frequency sampling and noise interference. To enable effective feature learning, a channel-wise spiking embedding layer is first applied to transform the input spike trains into a latent continuous representation. This embedded feature sequence is then processed by a series of spiking Transformer blocks, each leveraging the SSA mechanism to propagate and refine temporal information across the sequence. The overall transformation can be expressed as follows,
  \begin{equation}
    {S_{emb}}= SN(Linear(S))\label{eq:spiking_embedding}
  \end{equation}
  where $S_{emb}$ is the spiking embedding of the input spike trains
  $S$, and the linear layer is used to project the spike trains into a higher-dimensional
  space.

The spiking Transformer blocks consist of multiple stacked layers, each containing spiking self-attention and spiking feed-forward network modules. These components are designed to capture both local temporal features and long-range dependencies in the input spike sequences. The output sequence from the final spiking Transformer block is passed through a projection layer, such as a fully connected network, to map the learned spatiotemporal features to the final output of the BGFN, which is the predicted GPS position increment $\Delta P_{GPS}^{k{\rm{ + }}1}$. This method enables gradient computation across time steps and supports effective learning of dynamic motion patterns from sequential IMU data.

\section{Experimental Results and Analysis}
\subsection{Dataset tests}
\begin{figure}[!htp]
\centering
%   % Requires \usepackage{graphicx}
\includegraphics[width=.7\linewidth]{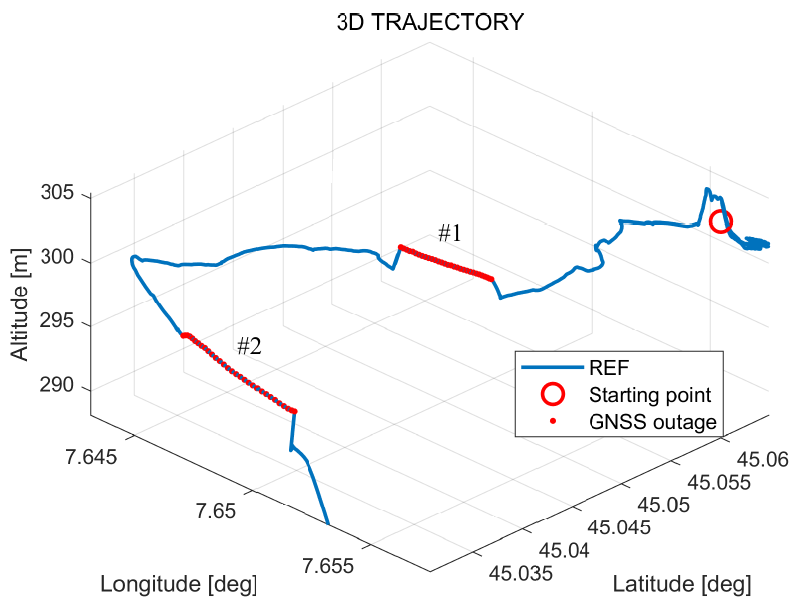}\\
\caption{The Navego dataset trajectory.}\label{3d}
\end{figure}

To evaluate the performance of the proposed algorithm, numerical experiments are conducted using a publicly available dataset from NaveGo \cite{gonzalez2015navego}, provided by Gonzalez and Dabove. The dataset is collected by driving a vehicle equipped with an Ekinox-D IMU and a GNSS receiver through the urban environment of Turin. Four different methods are compared: (1) standalone INS, (2) multilayer perceptron (MLP), (3) attention-based LSTM (AT-LSTM) \cite{chen2024error}, and (4) the proposed BGFN. Two 30-second GPS outages are selected from the dataset to assess the effectiveness of BGFN in aiding the INS during outages. The 3D trajectory of the NaveGo dataset is illustrated in Fig. \ref{3d}, with the positions of the two outages indicated by red lines.

The proposed method is implemented using PyTorch and SpikingJelly \cite{fang2023spikingjelly}. Training is performed for 1000 epochs using the BPTT algorithm with an NVIDIA RTX A6000 GPU and a batch size of 8. To enable the network to capture sufficient temporal dynamics, the BGFN is configured with a sliding time window of  $N=200$, meaning it uses the most recent 200 IMU measurements to predict the next GPS position increment. The embedding layer and the feed-forward network within the spiking Transformer both have a feature dimension of 256. The multi-head self-attention mechanism employs 6 attention heads, and the overall architecture consists of 4 spiking Transformer blocks. The output layer contains 2 neurons, corresponding to the two-dimensional GPS position increment $\Delta P_{GPS}$ in the local horizontal plane.

\begin{figure}[!htp]
\centering
%   % Requires \usepackage{graphicx}
\includegraphics[width=.9\linewidth]{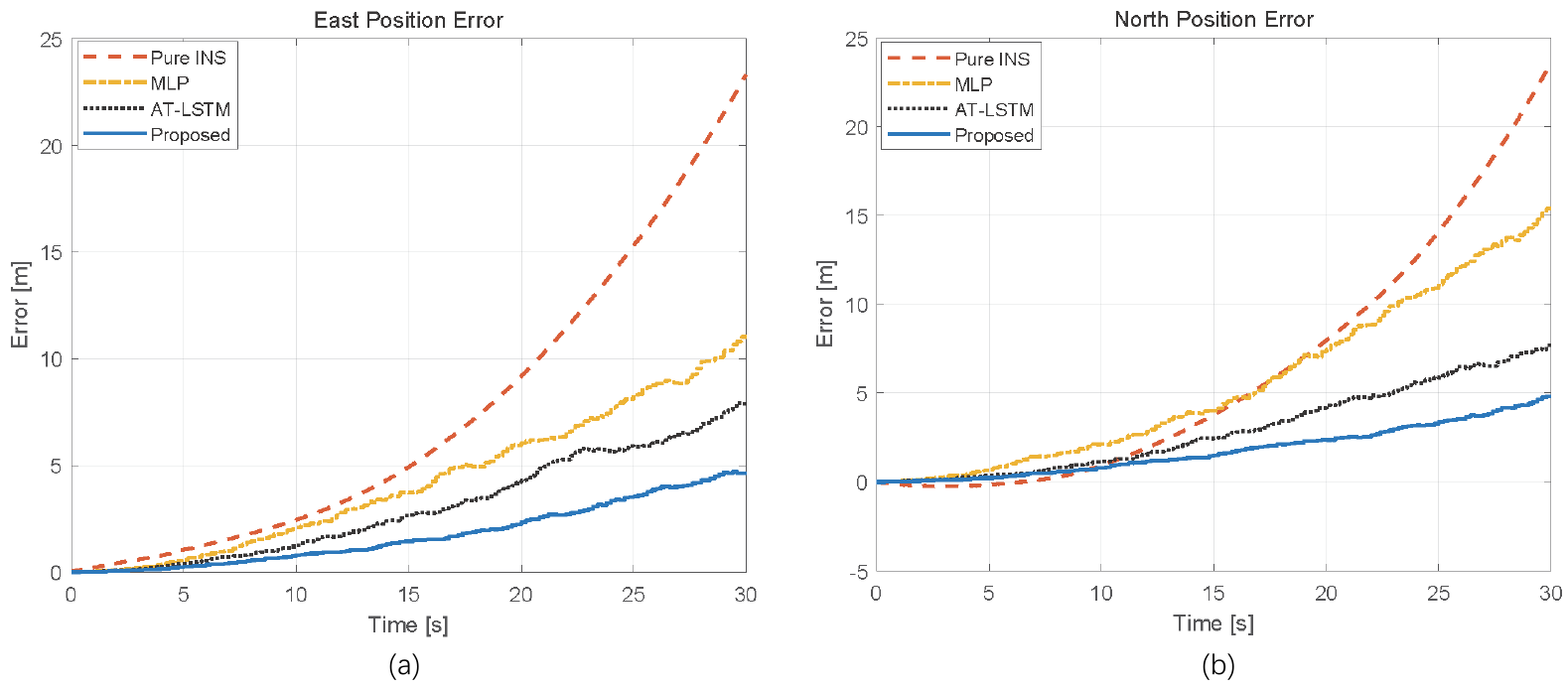}\\
\caption{ The position errors among different algorithms during the first outage in the dataset test. (a) The east position error. (b) The north position error.}\label{out1-exin}
\end{figure}

As illustrated in Fig. \ref{out1-exin} and Table \ref{tab-out1-exin}, the east and north position errors of four methods, namely pure INS, MLP, AT-LSTM, and BGFN, are compared. The results are shown using red, yellow, black, and blue curves, respectively. Due to the absence of external GPS corrections, the position errors of all methods gradually increase over time. The standalone INS exhibits the largest drift, with maximum east and north errors reaching 23.1 m and 23.2 m, respectively. The MLP model shows improved performance, reducing the maximum errors to 11.0 m in the east and 15.4 m in the north. The AT-LSTM further reduces the errors to 7.9 m and 7.7 m, demonstrating the benefit of modeling temporal dependencies with attention mechanisms. The proposed BGFN achieves the best performance among all methods, effectively suppressing error growth. Compared to standalone INS, the maximum errors are reduced by 79.7 $\%$ in the east and 79.3 $\%$ in the north, highlighting its superior capability in aiding inertial navigation during GPS outages.
\begin{table}[]
\centering
\caption{The max position error($m$) among different algorithms in the dataset test}\label{tab-out1-exin}
\scalebox{0.8}{
\begin{tabular}{ccccccccc}
\hline
\toprule[1pt]
\multirow{2}{*}{\textbf{Max Position Error (m)}} & \multicolumn{2}{c}{\textbf{INS}} & \multicolumn{2}{c}{\textbf{MLP}} & \multicolumn{2}{c}{\textbf{AT-LSTM}} & \multicolumn{2}{c}{\textbf{BGFN}} \\ \cline{2-9} 
                          & East         & North         & East       & North      & East       & North      & East        & North       \\ \hline
Outage 1                & 23.1         & 23.2        & 11.0       & 15.4      & 7.9       & 7.7       & \textbf{4.7}        & \textbf{4.8}        \\ 
Outage 2                & 10.6         & 19.6         & 5.6       & 9.3       & 2.9       & 6.0        & \textbf{1.8}        & \textbf{4.2}        \\ 
\toprule[1pt]
\end{tabular}
}
\end{table}

\begin{figure}[!htp]
\centering
%   % Requires \usepackage{graphicx}
\includegraphics[width=.9\linewidth]{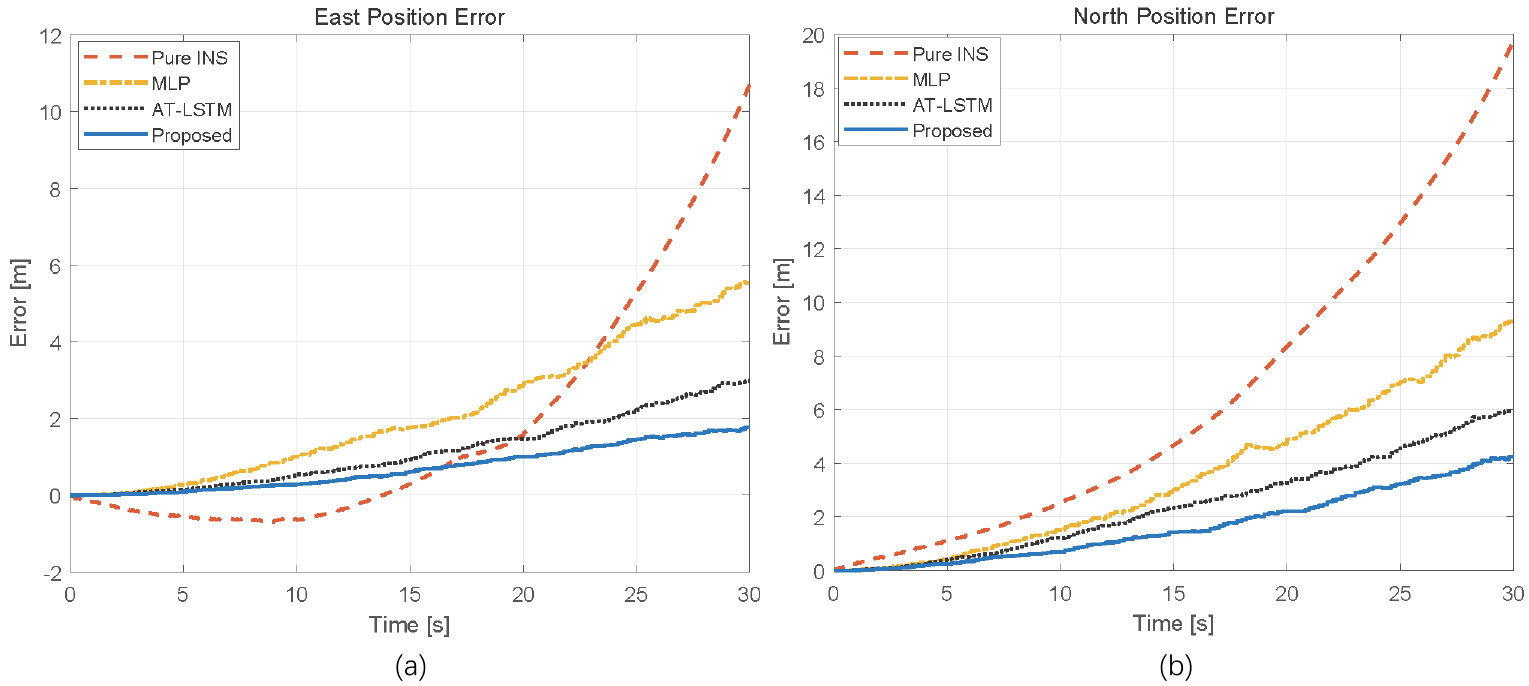}\\
\caption{  The position errors among different algorithms during the second outage in the dataset test. (a) The east position error. (b) The north position error.}\label{out2-exin}
\end{figure}

Fig. \ref{out2-exin} and Table \ref{tab-out1-exin} present the performance of the four methods during the second GPS outage. The results follow a similar trend to those observed in the first outage. The standalone INS exhibits the largest position drift, with maximum errors of 10.6 m in the east and 19.6 m in the north. Both MLP and AT-LSTM show improved accuracy compared to the standalone INS. Specifically, MLP achieves maximum errors of 5.6 m (east) and 9.3 m (north), while AT-LSTM reduces them to 2.9 m (east) and 6.0 m (north). The proposed BGFN demonstrates the best overall performance, with maximum errors further reduced to 1.8 m in the east and 4.2 m in the north. These results confirm that BGFN outperforms the pure INS, MLP, and AT-LSTM across multiple outage scenarios, demonstrating its superior ability to accurately predict GPS position increments and effectively assist inertial navigation under signal-denied conditions.

\subsection{Field tests}
\begin{figure}[!h]
  \centering
  \includegraphics[width=.8\linewidth]{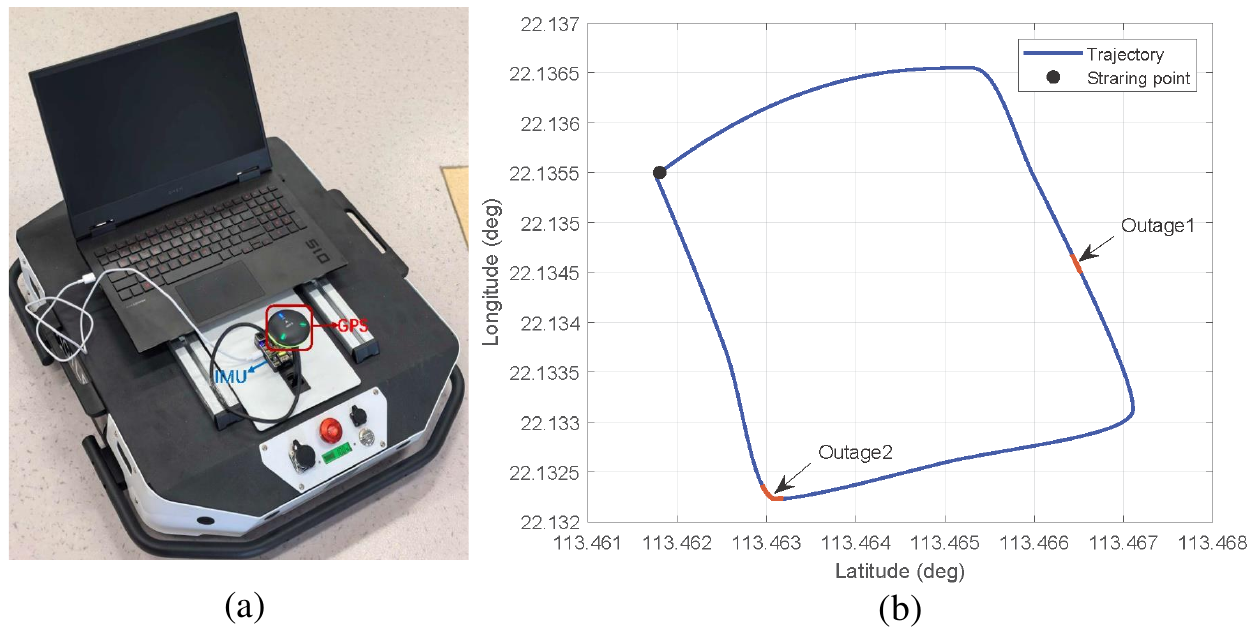}\\
  \caption{The mobile robot and trajectory. (a) The mobile vehicle. (b) The vehicle trajectory}\label{rob_tra}
\end{figure}

To further validate the proposed method, real-world field tests are conducted using a custom-built mobile car platform, as shown in Fig. \ref{rob_tra} (a). This platform is designed to collect realistic driving data under various environmental conditions and real-world sensor noise. The system integrates a Pixhawk autopilot, a high-performance embedded flight controller suitable for a wide range of robotic platforms, including fixed-wing aircraft, multi-rotor drones, helicopters, ground vehicles, and marine vessels. The primary inertial sensors are the Invensense MPU-6000 and STMicroelectronics LSM303D, providing gyroscope and accelerometer measurements with a typical gyroscope bias of 5°/s and accelerometer bias of 60 mg. To obtain accurate position reference data, a u-blox NEO-M9N GNSS receiver is installed. The position estimates are generated by fusing the GNSS and INS data using an EKF, providing a reliable ground truth trajectory for performance evaluation.

A mobile car driving experiment is conducted in Hengqin, Guangdong Province, China. The vehicle trajectory, shown in blue in Fig. \ref{rob_tra} (b), covers a realistic urban environment, and two 30-second GPS outages are indicated by red lines. To ensure reliable baseline navigation performance, the experiment is carried out in an open playground where GPS signals from at least seven satellites are consistently available throughout the normal operation periods. The GPS outages is artificially introduced using a signal shielding device to simulate signal-denied conditions, enabling a controlled evaluation of the proposed BGFN under realistic yet repeatable scenarios.
% \begin{figure}[!htp]
%   \centering
%   % Requires \usepackage{graphicx}
%   \includegraphics[width=.5\linewidth]{pic/robot.pdf}\\
%   \caption{Mobile car and integrated navigation system components.}\label{car}
% \end{figure}

\begin{figure}[!htp]
\centering
%   % Requires \usepackage{graphicx}
\includegraphics[width=.9\linewidth]{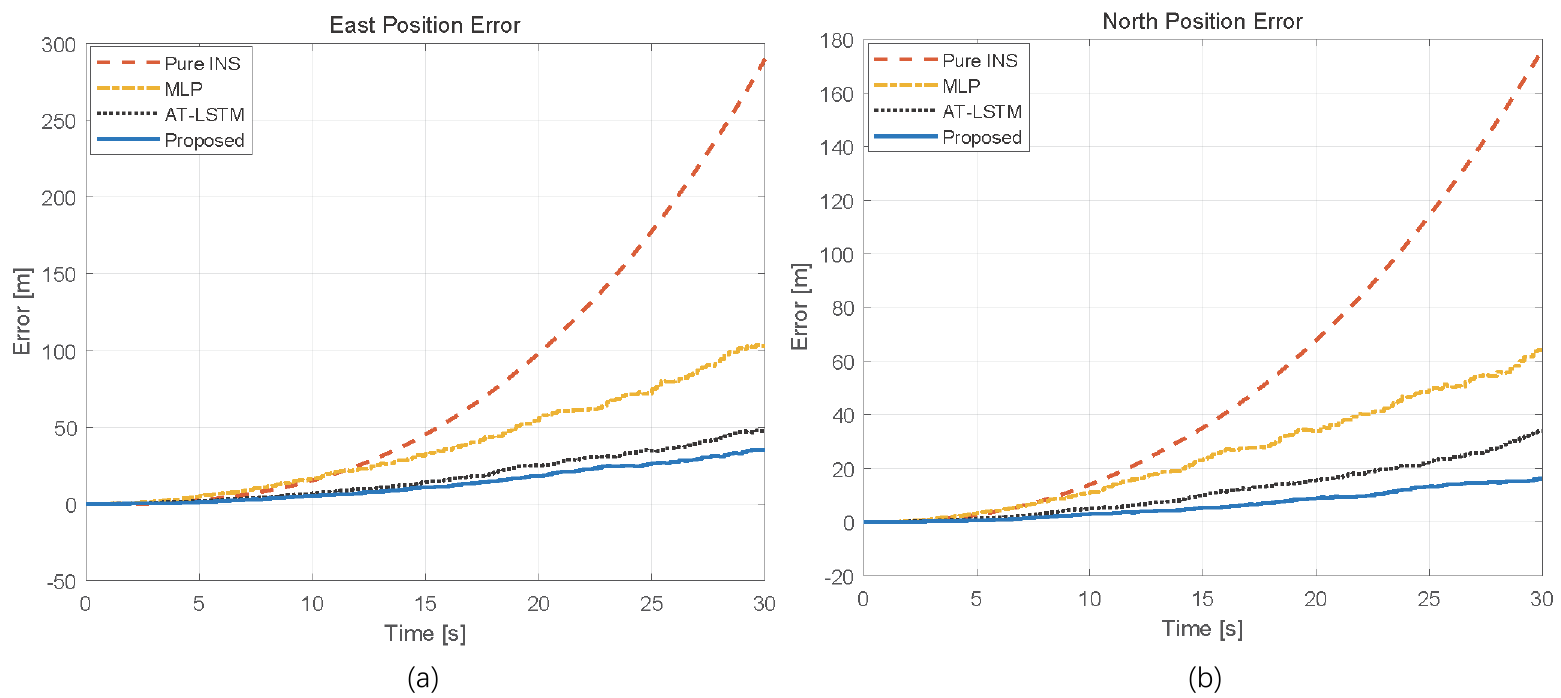}\\
\caption{ The position errors among different algorithms during the first outage in the field test. (a) The east position error. (b) The north position error}\label{out1-mpu}
\end{figure}

Fig. \ref{out1-mpu} illustrates the east and north position errors estimated by different algorithms during the first GPS outage, with the maximum errors in both directions summarized in Table \ref{tab-real}. When GPS signals are interrupted, the KF loses external position updates, causing the integrated navigation system to operate in standalone INS mode. In this mode, position errors accumulate rapidly over time due to unbounded drift from sensor biases and noise. When MLP, AT-LSTM, or the proposed BGFN is employed to aid the INS, the error growth is mitigated to varying degrees. During the first outage, the maximum east position errors for standalone INS, MLP, AT-LSTM, and BGFN are 289.4 m, 103.0 m, 47.3 m, and 35.0 m, respectively, while the corresponding north errors are 175.1 m, 64.2 m, 33.8 m, and 16.2 m. Comparative analysis shows that the BGFN-based method achieves the smallest position errors among all approaches. These results demonstrate that the BGFN effectively enhances navigation accuracy during GPS outages and exhibits superior capability in suppressing inertial navigation drift compared to traditional deep learning models such as MLP and AT-LSTM.

\begin{figure}[!htp]
\centering
%   % Requires \usepackage{graphicx}
\includegraphics[width=.9\linewidth]{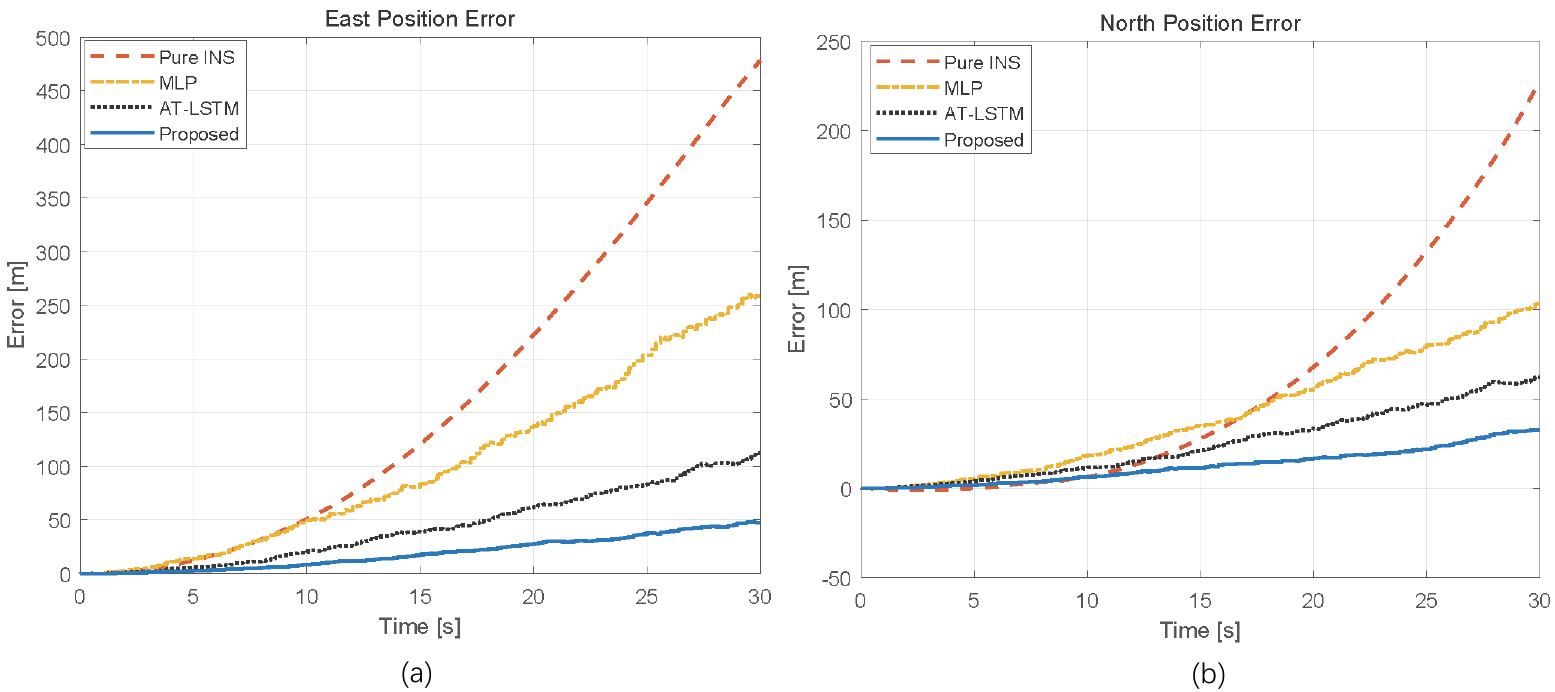}\\
\caption{ The position errors among different algorithms during the second outage in the field test. (a) The east position error. (b) The north position error.}\label{out2-mpu}
\end{figure}

To further evaluate the performance of BGFN-assisted integrated navigation under dynamic motion conditions, the second GPS outage is selected during a vehicle maneuvering phase characterized by significant changes in direction. As shown in Fig. \ref{out2-mpu}, the east and north position errors of the standalone INS, MLP, AT-LSTM, and BGFN methods are compared during the second outage period, with quantitative results summarized in Table \ref{tab-real}. At the end of the outage, the BGFN-based method reduces the position error by 90.0 $\%$ in the east and 85.5 $\%$in the north compared to standalone INS. These results demonstrate that the proposed BGFN maintains high prediction accuracy even during large-scale maneuvers, effectively constraining inertial drift and significantly improving navigation robustness under challenging, dynamic conditions.

The results from the mobile robot real-world driving experiments demonstrate that the BGFN-based integrated navigation method effectively suppresses position error growth and achieves higher navigation accuracy compared to traditional neural network approaches. Furthermore, when deployed on a 45 nm neuromorphic hardware platform \cite{horowitz20141}, the proposed spiking Transformer architecture achieves a theoretical energy reduction of 66.3 $\%$ relative to conventional Transformer-based models \cite{lv2024efficient}. This significant improvement underscores the energy efficiency of the bio-inspired design, making it well suited for edge computing applications where power consumption is a critical constraint.
\begin{table}[]
\centering
\caption{The max position error  among different algorithms in the real field test}\label{tab-real}
\scalebox{0.8}{
\begin{tabular}{ccccccccc}
\hline
\toprule[1pt]
\multirow{2}{*}{\textbf{Max Position Error (m)}} & \multicolumn{2}{c}{\textbf{INS}} & \multicolumn{2}{c}{\textbf{MLP}} & \multicolumn{2}{c}{\textbf{AT-LSTM}} & \multicolumn{2}{c}{\textbf{BGFN}} \\ \cline{2-9} 
                          & East         & North         & East       & North      & East       & North       & East        & North       \\ \hline
Outage 1                & 289.4         & 175.1         & 103.0       & 64.2       & 47.3       & 33.8        & \textbf{35.0}        & \textbf{16.2}        \\ 
Outage 2                 & 477.5         & 226.0         & 258.7       & 103.0       & 112.3       & 62.1        & \textbf{47.6}        & \textbf{32.7}        \\ 
\toprule[1pt]
\end{tabular}
}
\end{table}

\section{Conclusion}
This paper proposes a novel spiking neural network-based method for aiding INS during GPS signal outages by effectively mitigating the accumulation of navigation errors. The key advantage of the proposed BGFN lies in its ability to not only extract robust feature representations from noisy sensor measurements but also to automatically correlate current inputs with historical model states, enabling accurate temporal modeling. To evaluate its performance, both experiments on numerical public datasets and real-world field tests are conducted, and the results demonstrate that BGFN significantly improves navigation accuracy under GPS-denied conditions. The proposed brain-inspired GPS/INS fusion architecture is capable of capturing the inherent nonlinear relationship between INS outputs and GPS position increments, thereby providing reliable and precise navigation solutions during prolonged signal outages. Although the method shows strong performance, there remains room for improvement: future work will focus on deploying BGFN onto neuromorphic or brain-inspired hardware platforms to leverage its energy efficiency, as well as exploring an end-to-end trainable spiking neural network for the GPS/INS integrated system to further enhance accuracy and robustness in complex environments.

 \bibliographystyle{elsarticle-num} 
 \bibliography{cas-refs}
 
 \section*{Acknowledgments}
 This work is supported by the Science and Technology Development Fund, Macau, SAR, under Grant 0005/2023/EIB1.

\end{document}